\documentclass[]{bytedance_seed}

\usepackage[toc,page,header]{appendix}

\usepackage{minitoc}
\usepackage{solarized-light}
\usepackage{multirow}
\usepackage{graphicx}
\usepackage{array}
\usepackage{makecell}
\usepackage{framed}
\usepackage{hyperref}
\usepackage{amsmath} 
\usepackage[colorinlistoftodos]{todonotes}
\usepackage{longtable}
\usepackage{hhline}
\usepackage{fancyvrb}
\usepackage{fvextra}
\usepackage{CJKutf8}
\usepackage{multicol}
\usepackage{cleveref}
\usepackage{tablefootnote}
\usepackage{threeparttable}
\usepackage{tabularx}
\usepackage{mdframed}
\usepackage{subcaption}
\usepackage{amssymb}
\usepackage[usestackEOL]{stackengine}
\usepackage[numbers]{natbib}
\newcommand{\commentout}[1]{}
\renewcommand{\paragraph}[1]{\noindent\textbf{#1.}\hspace*{1em}}
\usepackage{enumitem}
\setlist[itemize]{leftmargin=15pt}

\RequirePackage{xspace}
\makeatletter
\DeclareRobustCommand\onedot{\futurelet\@let@token\@onedot}
\def\@onedot{\ifx\@let@token.\else.\null\fi\xspace}

\makeatother

\title{Timer-S1: A Billion-Scale Time Series Foundation Model with Serial Scaling}

\author[1,2,*]{Yong Liu}
\author[1,2,*]{Xingjian Su}
\author[2,*]{Shiyu Wang}
\author[1,*]{Haoran Zhang}
\author[1]{Haixuan Liu}
\author[1]{Yuxuan Wang}
\author[2]{Zhou Ye}
\author[2]{Yang Xiang}
\author[1]{Jianmin Wang}
\author[1, \dagger]{Mingsheng Long}
\contribution[*]{Equal contribution}

\affiliation[1]{Tsinghua University}
\affiliation[2]{ByteDance}
\contribution[\dagger]{Corresponding author}

\abstract{We introduce \textbf{Timer-S1}, a strong Mixture-of-Experts (MoE) time series foundation model with 8.3B total parameters, 0.75B activated parameters for each token, and a context length of 11.5K. To overcome the scalability bottleneck in existing pre-trained time series foundation models, we perform \textit{Serial Scaling} in three dimensions: model architecture, dataset, and training pipeline. Timer-S1 integrates sparse TimeMoE blocks and generic TimeSTP blocks for \textit{Serial-Token Prediction (STP)}, a generic training objective that adheres to the \textit{serial nature of forecasting}. The proposed paradigm introduces serial computations to improve long-term predictions while avoiding costly rolling-style inference and pronounced error accumulation in the standard next-token prediction. Pursuing a high-quality and unbiased training dataset, we curate TimeBench, a corpus with one trillion time points, and apply meticulous data augmentation to mitigate predictive bias. We further pioneer a post-training stage, including continued pre-training and long-context extension, to enhance short-term and long-context performance. Evaluated on the large-scale GIFT-Eval leaderboard, Timer-S1 achieves state-of-the-art forecasting performance, attaining the best MASE and CRPS scores as a pre-trained model. Timer-S1 is released to facilitate further research.}

\date{\today}
\correspondence{Mingsheng Long at \email{mingsheng@tsinghua.edu.cn}}
\checkdata[Model]{\url{https://huggingface.co/bytedance-research/Timer-S1}}

\begin{document}
\maketitle

\newpage
\tableofcontents
\newpage

\section{Introduction}
Time series serve as a fundamental resource across a wide spectrum of real-world applications~\cite{friedman1962interpolation, box2013box, hyndman2018forecasting, breunig2000lof, kendall1953analysis}, ranging from industrial sensing and financial assessment to healthcare monitoring and climate forecasting. Its inherent capacity to capture dynamic processes and evolving patterns positions it as one of the pivotal modalities on the path toward Artificial General Intelligence (AGI). In recent years, pre-trained \textbf{time series foundation models} have advanced rapidly~\cite{das2023decoder, woo2024unified, goswami2024moment, ansari2024chronos, shi2024time}, progressively narrowing the gap between model development and deployment~\cite{wu2023interpretable, tu2024powerpm, zhang2024trajectory, zhu2025fincast}. The emergence of time series foundation models has initiated a paradigm shift toward \textit{General Forecasting}, serving as universal forecasters that are free from task-specific training~\cite{das2023decoder, cohen2024toto, liu2025sundial, ansari2025chronos, auer2025tirex} and enabling agentic systems to reason about time series data~\cite{zhao2025timeseriesscientist, garza2025timecopilot, das2025synapse}.

Building capable time series foundation models is particularly challenging due to the inherent complexity of time series data. Unlike natural language, which follows grammatical rules and common knowledge~\cite{pinker2003language}, or images and videos that often lie on low-dimensional manifolds~\cite{chapelle2009semi}, time series are characterized by significant distributional heterogeneity across domains. Besides, its variability in frequencies and shapes magnifies the challenge of capturing multi-scale dependencies and interactions among multivariate signals~\cite{box2013box}, which reside in high-dimensional and unstructured raw data. Furthermore, the inherent non-stationary and stochastic nature of real-world processes, where temporal dynamics can shift unexpectedly due to external factors or regime changes, introduces substantial uncertainty in forecasting~\cite{priestley1988non}.

To address the aforementioned challenges, we have devoted consistent efforts toward unified and capable time series foundation models. Our initial work, Timer~\cite{liutimer}, coped with the domain heterogeneity by leveraging a general-purpose decoder-only Transformer~\cite{vaswani2017attention} and a next-patch prediction objective, which enables unified pre-training across different datasets. Its successor, Timer-XL~\cite{liu2024timer}, addressed structural variability through a generic self-attention mechanism capable of modeling multi-dimensional time series in an autoregressive approach. More recently, Timer-3~\cite{liu2025sundial} (Sundial), equipped with an accelerated Transformer backbone and multi-patch prediction objective, introduced the generative forecasting paradigm based on flow matching~\cite{lipman2022flow} to tackle the inherent uncertainty in time series forecasting. While these prior explorations established feasible groundworks, they operated primarily within limited model sizes. This constraint brings us to the central goal of this technical report: \textbf{effectively scaling} a time series foundation model that delivers substantially stronger forecasting performance while reducing inference cost.

\begin{figure*}[htb]
\centering
\includegraphics[width=\textwidth]{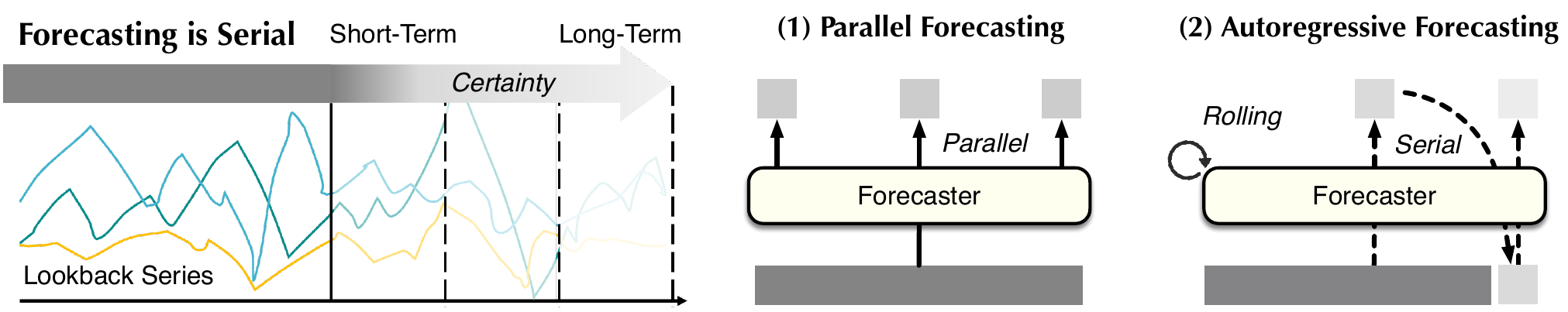}
\caption{Forecasting into the long term accumulates uncertainty, as the prediction of each step depends on all preceding estimations, which positions time series forecasting as a serial problem~\cite{liu2025serial}. Parallel-forecasting models, which predict multiple future steps simultaneously, do not scale with sufficient serial computations to reliably capture the recurrent dependencies. Although autoregressive models mirror the serial nature of the task by "predicting step by step", their iterative rolling mechanism over the input still entails significant computational overhead.} 
\label{fig:motivation}
\end{figure*}

\begin{figure*}[tb]
\centering
\includegraphics[width=\textwidth]{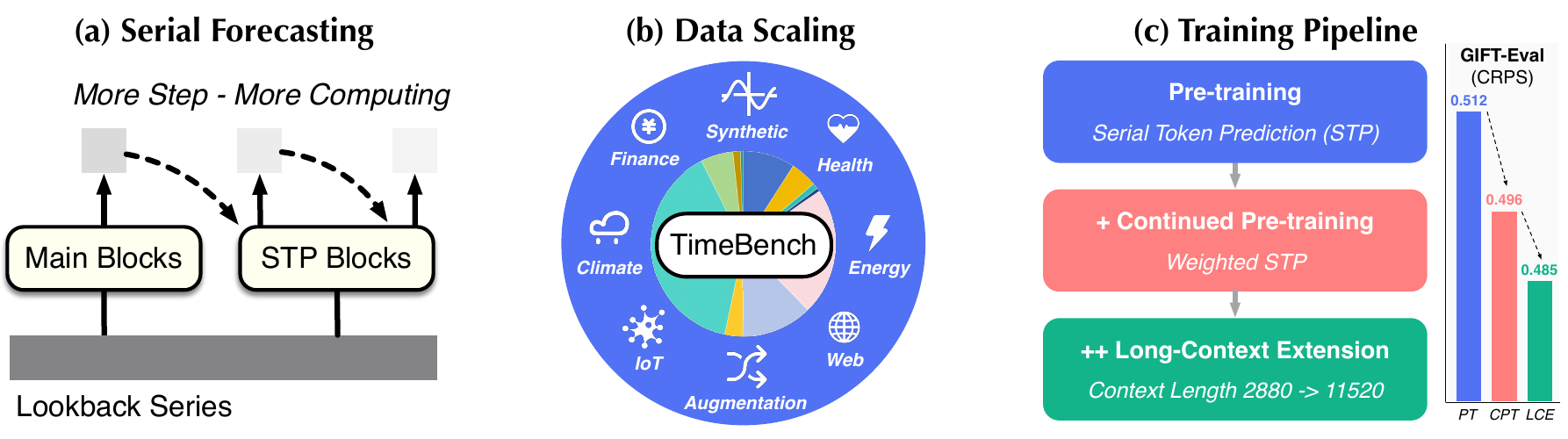}
\caption{The Serial Scaling of Timer-S1 is achieved by (a) \textit{serial forecasting}, which efficiently produces multi-step prediction with serial computations; (b) \textit{data scaling} with data augmentation applied to TimeBench~\cite{liu2025sundial}, a corpus of over one trillion time points; and (c) \textit{post-training} that comprehensively enhances the capability of the model.} 
\label{fig:serial_scaling}
\end{figure*}

To advance the scalability of time series foundation models, we introduce \textbf{Timer-S1}, a sparse Mixture-of-Experts (MoE) model with 8.3B parameters, of which only 0.75B are activated for each token. Recognizing the \textit{serial nature} of time series forecasting (Figure~\ref{fig:motivation}), Timer-S1 leverages an efficient \textit{serial forecasting} approach. It incorporates essential serial computations missing in parallel forecasting, while avoiding redundant rolling operations in autoregressive models. Technically, serial forecasting is implemented through a generic TimeSTP block for \textit{Serial-Token Prediction (STP)}, a serialized version of the Transformer block. As shown in Figure~\ref{fig:serial_scaling} (a), each TimeSTP block refers to the initial lookback series and intermediate representations, and iteratively produces the shift-by-one prediction, thereby introducing progressive serial computations for multi-horizon forecasts. Crucially, these TimeSTP blocks are retained during inference, allowing the model to adaptively generate predictions without rolling-style autoregression. Consequently, Timer-S1 has a flexible context length as an autoregressive model while producing multi-step predictions in a single forward pass.

Beyond architectural design, the serial scaling of Timer-S1 includes data scaling and post-training strategy. As illustrated in Figure~\ref{fig:serial_scaling}, Timer-S1 is first pre-trained on TimeBench, a dataset comprising over one trillion time series points. To mitigate predictive bias, we employ data augmentation techniques including resampling and value‑flipping. During the pre-training stage, the model is densely supervised, where a training sample becomes multiple forecasting tasks with variable input and output lengths. Next, Timer-S1 is continuously pre-trained using a weighted STP objective that decays with the forecasting horizon, which aims to improve the short-term performance. During the post-training stage, we also scale the context window based on the RoPE~\cite{su2024roformer} implementation, enhancing the model's capability to handle longer sequences. By decoupling the training pipeline, we lower the overall training cost and encourage different representation learning objectives, improving the overall forecasting performance on long-term and short-term tasks.

Timer-S1 achieves state-of-the-art performance metrics on the GIFT-Eval~\cite{aksugift} benchmark (CRPS: 0.485, MASE: 0.693). Our analysis reveals that the proposed serial forecasting paradigm delivers particularly strong gains on medium- and long-term horizons, delivering a generic idea to scale up existing time series foundation models. Beyond leaderboard results, we conduct a series of analytical experiments: (1) an exploration of the scaling law by identifying the optimal model configuration, (2) a comparison of serial-token prediction against standard training objectives such as next-token prediction and multi-token prediction~\cite{das2023decoder, liu2024deepseek}, and (3) ablation studies on key components of Timer-S1, including modular design of TimeSTP, data augmentation, and pre-training benefits. Together, these findings validate the General Forecasting capability of Timer-S1 and provide a possible approach for advancing time series foundation models.

The remainder of this paper is organized as follows. Section~\ref{sec:background} reviews the background of time series foundation models and discusses the design motivation in Timer-S1. Section~\ref{sec:arch} introduces the architecture of Timer-S1 and its implementation for serial forecasting. Section~\ref{sec:train} details the training pipeline, including the dataset, pre-training, post-training, and training infrastructure. Section~\ref{sec:exp} presents evaluation results, scaling analysis, and ablation studies. Finally, Section~\ref{sec:conclusion} concludes the work, discusses limitations, and suggests future directions.

\section{Background}\label{sec:background}
Time series forecasting is a fundamental and ubiquitous task of predicting future values based on observed historical sequences~\cite{kendall1953analysis}. This task is widely studied across scientific and industrial fields. Methodologically, the field has evolved through four phases: early reliance on statistical methods~\cite {box2013box}, such as ARIMA and Exponential Smoothing, which furnish solid theoretical foundations but may struggle with complex nonlinear patterns; followed by machine learning models~\cite{hyndman2018forecasting} (e.g., SVR, tree-based models) that facilitate data-driven approaches with increased robustness. In recent years, deep learning models~\cite{lim2021time, wen2022transformers, wang2024deep} (e.g., TCNs, RNNs, Transformers) have achieved breakthroughs by leveraging their strong feature extraction and sequence modeling capabilities on ever larger data. Nevertheless, these models are typically trained from scratch on specific tasks, leading to poor generalization in cold-start and data-scarce scenarios. This limitation has spurred the emergence of time series foundation models, which aim to learn various evolving patterns via large-scale pre-training and subsequently adapt to downstream tasks, embodying a "train once, apply anywhere" paradigm shift.

\begin{figure*}[tb]
\centering
\includegraphics[width=\textwidth]{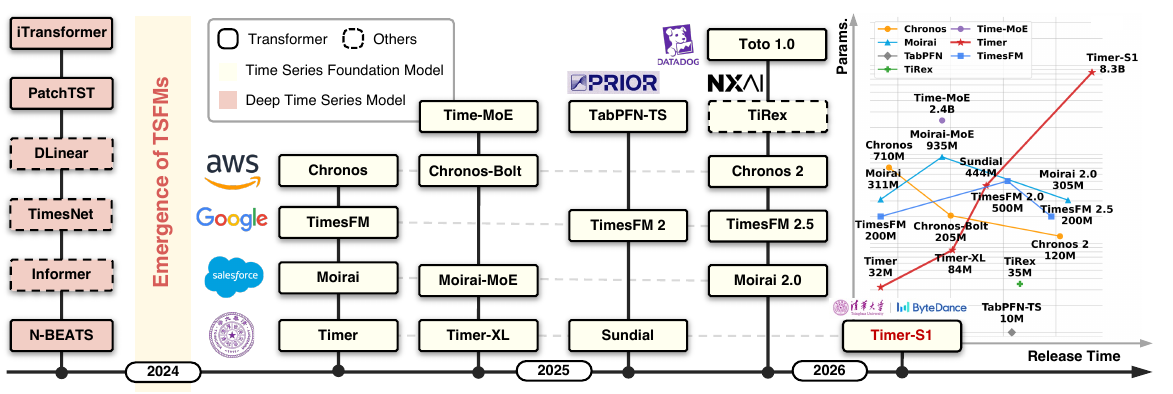}
\caption{A timeline of representative time series forecasting models in recent years. This timeline is established according to the release date of the paper or technical report for a model. Notably, the Timer model is a continuously developed family of time series foundation models that has presents sustained scaling in model size across its generations.} 
\label{fig:related_work}
\end{figure*}

Recent research on time series foundation models has focused on addressing several foundational challenges, yielding substantial progress including (1) specialized architectures, developing neural network structures capable of effectively capturing multi-scale and multi-dimensional dependencies~\cite{woo2024unified, liu2024timer, cohen2024toto, ansari2025chronos, auer2025tirex}; (2) pre-training paradigms, such as data normalization~\cite{kim2021reversible}, training objective~\cite{das2023decoder, woo2024unified}, and loss functions~\cite{ansari2024chronos, liu2025sundial} tailored for time series; (3) data curation, integrating data governance, selection and synthesis techniques~\cite{dooley2023forecastpfn, liutimer, woo2024unified, ansari2024chronos}, (4) model adaptation, leveraging pre-trained foundation models on specific tasks or incorporating exogenous factors~\cite{arango2025chronosx, das2024context, qin2025cora}, and (5) agentic tools, integrating into agent-based AI systems, e.g., large language models, to build an autonomous forecasting pipeline~\cite{zhao2025timeseriesscientist, garza2025timecopilot, das2025synapse}. 

As time series foundation models have attracted increasing attention, a critical "scaling bottleneck" became apparent (Figure~\ref{fig:related_work}). While mature scalable architectures, such as Mixture-of-Experts~\cite{jacobs1991adaptive, vaswani2017attention}, have been well-established in large language models, prior attempts to adapt them to time series~\cite{shi2024scaling, liu2024moirai} may lead to inferior performance or fail to achieve a significant breakthrough in model scale. In this technical report, we note that the effectiveness of scaling fundamentally relies on respecting the nature of the forecasting task, that is, time series forecasting is an inherently serial problem~\cite{liu2025serial}, where long-horizon accuracy depends on progressive step-by-step reasoning.  Notably, autoregressive models naturally satisfy the serial natural and have been proven to be scalable in our previous work~\cite{liutimer}, corroborated by the fact that many time series foundation models adopt decoder-only Transformers~\cite{vaswani2017attention} and next-token prediction~\cite{bengio2000neural} for pre-training. 

However, the rolling steps required for long-term autoregressive forecasting incur substantial computational overhead and error accumulation. To address this issue, multi-token prediction initially adopted in LLMs~\cite{stern2018blockwise, gloeckle2024better} has been introduced to time series foundation models~\cite{das2023decoder, liu2025sundial, liu2025moirai}. Nonetheless, such adaptations are mainly designed from a representation learning perspective, i.e., the backbone is forced to extract shared representations for both long- and short-term forecasting, which poses optimization difficulties and can still lead to a scaling bottleneck. Therefore, we propose Serial-Token Prediction (STP), which conducts adaptive serial computations at different forecasting horizons. Besides, while accelerating standard autoregressive forecasting with KV-Cache~\cite{pope2023efficiently} can preserve serial computations and reduce inference costs, complex evolution patterns inherent in time series make error accumulation particularly pronounced. Hence, the proven methodologies from LLMs cannot be directly applied to time series foundation models. Technically, we retained TimeSTP blocks after pre-training to minimize this train–test gap, which departs from the methodology in LLMs~\cite{stern2018blockwise, gloeckle2024better, liu2024deepseek}.

In addition to architectural design, this technical report is dedicated to systematically enhancing the usability of time series foundation models. First, by overcoming the scaling bottleneck, we pre-train Timer-S1 on a massive volume of real-world and synthetic data, an augmented trillion-scale time series corpora to help the model recognize general evolving patterns in time series. Second, Timer-S1 adopts a continued pre-training (CPT) strategy for time series foundation models. Different from fine-tuning~\cite{arango2025chronosx, benechehab2025adapts, qin2025cora}, which typically focuses on adapting to a narrow downstream dataset, CPT is designed to enhance the model for a particular type of capability (e.g., short-term forecasting). In contrast, single-stage pre-training may lead to training difficulties, because pre-training on unified datasets may overlook the task discrepancy, i.e., short-term and long-term forecasting tasks require different training objectives and training data.

\section{Timer-S1: Architecture}\label{sec:arch}
\begin{figure*}[htb]
\centering
\includegraphics[width=\textwidth]{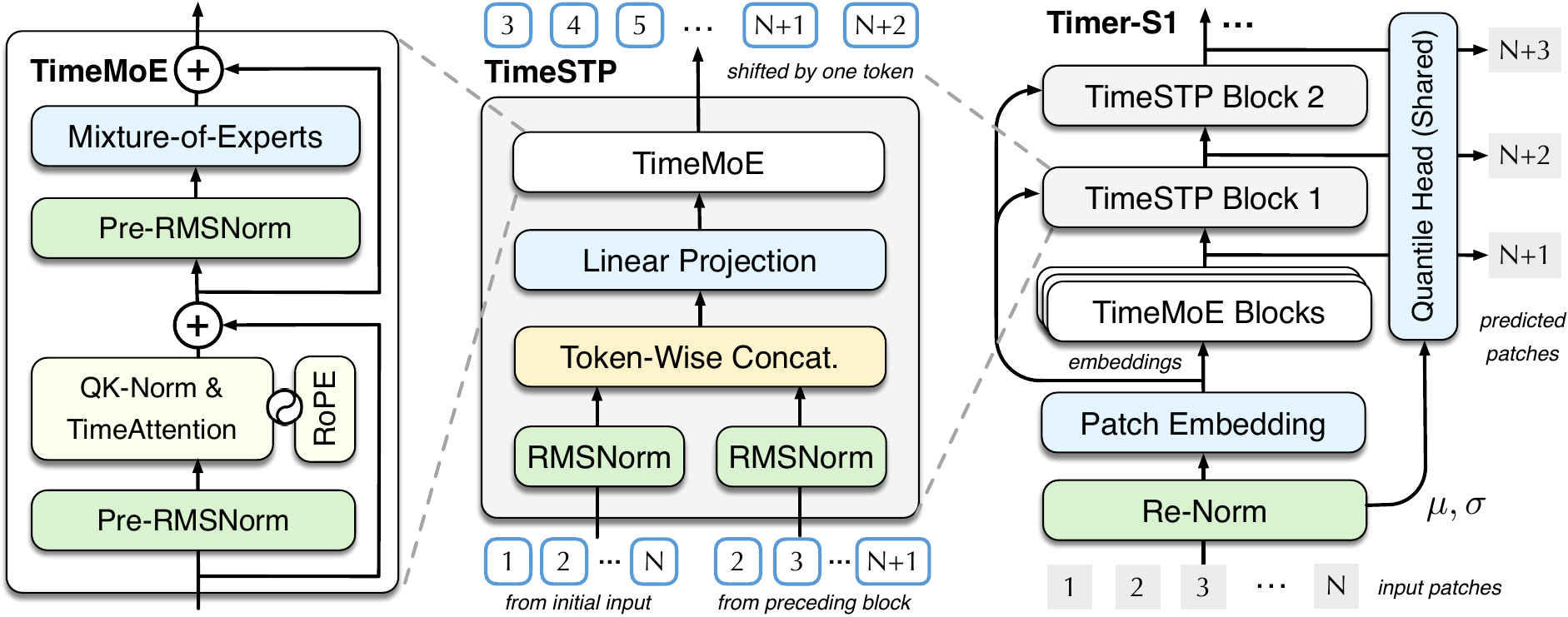}
\caption{Overall architecture of Timer-S1. The input time series is re-normalized and divided into patch tokens. These patch embeddings are fed into a decoder-only Transformer. The Transformer backbone consists of a series of TimeMoE blocks, where Pre-RMSNorm and QK-Norm~\cite{henry2020query} are adapted for training stability, followed by a sequence of TimeSTP blocks. TimeSTP extends TimeMoE by additionally conditioning on the initial input embeddings, iteratively refining the token embeddings from the previous block, and generating shifted-by-one token predictions. All output embeddings are projected by a shared forecasting head to produce quantile predictions. Timer-S1 enables serial forecasting, where predictions of longer horizons actually undergo more serial computations in the Transformer block.} 
\label{fig:arch}
\end{figure*}

In this section, we introduce the architecture of Timer-S1. As depicted in Figure~\ref{fig:arch}, Timer-S1 consists of three parts: (1) an instance re-normalization to address value discrepancy and a patch-level token embedding, (2) a decoder-only Transformer backbone enhanced with specialized TimeMoE blocks to tackle data heterogeneity, and TimeSTP blocks for serial forecasting, and (3) a shared forecasting head trained with a quantile loss tailored for evaluation on the GIFT-Eval leaderboard.

\subsection{Normalization and Embedding}
Time series foundation models are typically pre-trained on diverse datasets with varying scales, input shapes, and variate semantics, which complicates the direct application of a standardized 1D next-token prediction. To address this, the pre-training of Timer-S1 is designed to capture univariate evolving patterns, that is, to capture temporal dependencies from historical context and produce the most plausible future trends.

As shown in Figure~\ref{fig:training}, we adopt a single-series sequence format~\cite{liutimer}, where multivariate time series are split into univariate series and normalized per instance. This approach eliminates semantics and correlations of variates, as such features may be dataset-specific and unstable in cross-domain generalization. To build a general-purpose pre-trained model, Timer-S1 is therefore oriented toward acquiring a foundational forecasting capability, grounded primarily on in-context univariate patterns. To compensate for the loss of multivariate structures, variate-level configurations can be later incorporated via task-specific fine-tuning.

\paragraph{Re-Normalization} Given a univariate time series input $\{x_1, \dots, x_{T}\}$ and future time series $\{x_{T+1}, \dots, x_{T+F}\}$, Timer-S1 conducts instance re-normalization to mitigate scale discrepancy as follows:
\begin{equation}
\mu = \frac{1}{T}\sum_{t=1}^T x_t,\ \sigma^2 =\frac{1}{T} \sum_{t=1}^T (x_t-\mu)^2,\ \tilde x_t = \frac{x_t-\mu}{\sigma},\ t=1,\dots,T,
\end{equation}
where $\mu$ and $\sigma$ are the mean and standard deviation of the input, and will be reused for de-normalizing for the model's final outputs $\hat{x}_i$, thereby restoring the original data scale as follows:
\begin{equation}\label{equ:denorm}
\hat{x}_t=\sigma\cdot \tilde{x}_t + \mu,\ t=T+1,\cdots,T+F.
\end{equation}
As a result, value-shifting and scaling on the input are equally propagated in the future predictions, making the model robust on varying scales and allowing it to concentrate on learning local temporal patterns.

\paragraph{Patch Embedding} Timer-S1 adopts patch tokenization~\cite{nie2022time}. A patch token is defined as consecutive time points with length $P$. The $i$-th patch is denoted as $\tilde {\mathbf{x}}_i=\{\tilde x_{1+(i-1)P}, \dots, \tilde x_{iP}\}$. The normalized input is divided into $N=\lceil T/P \rceil$ patches. For non-divisible length, we conduct left-padding and employ a binary mask $\mathbf{m}_t\in\mathbb{R}^P$ per patch to indicate padded positions. Next, a residual network$: \mathbb{R}^{2P} \mapsto \mathbb{R}^D$ embeds all tokens:
\begin{equation}
\mathbf{h}_i^0 = \operatorname{PatchEmbed}\big(\operatorname{Concat}(\tilde {\mathbf{x}}_i, \mathbf{m}_i)\big),\ i=1,\dots,N,
\end{equation}
where $\mathbf{h}_i^0\in\mathbb{R}^D$ denotes initial input embedding and $D$ is the hidden dimension of the Transformer.

\subsection{Transformer Backbone}
Timer-S1 is a decoder-only Transformer that inherits the core design of Timer. As shown in Figure~\ref{fig:arch}, Timer-S1 includes a stack of TimeMoE blocks (main blocks), followed by a series of TimeSTP blocks, which incorporate a TimeMoE module and introduce serial computations required for serial-token prediction.

\paragraph{TimeMoE} To address the heterogeneity of time series data, we adopt sparse Mixture-of-Experts (MoE) to adaptively assign experts to process time series patches with distinct patterns. Each TimeMoE block consists of a Multi-Head Attention (MHA) module and a MoE module. In order to improve training stability, each block (indexed by $l$) incorporates Pre-RMSNorm~\cite{zhang2019root, xiong2020layer}, formulated as:
\begin{equation}
\begin{aligned}
\mathbf{u}_i^l & =\operatorname{MHA}\left(\operatorname{RMSNorm}(\mathbf{h}_i^{l-1})\right)+\mathbf{h}_i^{l-1},\\
\mathbf{h}_i^l & =\operatorname{MoE}\left(\operatorname{RMSNorm}(\mathbf{u}_i^l)\right)+\mathbf{u}_i^l .
\end{aligned}
\label{equ:timemoe}
\end{equation}
For a single-head attention module, we adopt QK-Norm~\cite{henry2020query} with $\ell_2$ normalization to mitigate the saturation of softmax scores, a known issue identified in the prior work~\cite{liu2022non} (the block index $l$ is omitted for simplicity):
\begin{equation}
\hat{\mathbf{q}}_i=\frac{\mathbf{W}_{\mathbf{q}}^\top\mathbf{h}_i}{||\mathbf{W}_{\mathbf{q}}^\top\mathbf{h}_i||},\ \hat{\mathbf{k}}_i =\frac{\mathbf{W}^\top_{\mathbf{k}}\mathbf{h}_i}{||\mathbf{W}^\top_{\mathbf{k}}\mathbf{h}_i||},
\end{equation}
where $\mathbf{W}_\mathbf{q}, \mathbf{W}_\mathbf{k}, \mathbf{W}_\mathbf{v} \in \mathbb{R}^{D\times d}$ project embeddings $\mathbf{H}=\{\mathbf{h}_1,\dots ,\mathbf{h}_N\}\in\mathbb{R}^{N\times D}$ into $d$-dimensional queries, keys, and values, respectively. Subsequently, a causal self-attention mechanism with Rotary Position Embedding (RoPE)~\cite{su2024roformer} is applied to encode the relative position of each patch token, which is formulated as follows:
\begin{equation}
\begin{aligned}
\mathcal{A}_{i,j} &= \hat{\mathbf{q}}^\top_i\mathbf{R}_{\Theta, i-j}\hat{\mathbf{k}}_j,\ i, j = 1, \dots, N,\\
\operatorname{Attention}(\mathbf{H}) &= \operatorname{Softmax}\left(\tau * \operatorname{Mask}(\mathcal{A})\right) \mathbf{H}\mathbf{W}_\mathbf{v},
\end{aligned}
\end{equation}
where $\mathbf{R}_{\Theta, t}\in\mathbb{R}^{d\times d}$ is a rotary matrix with a rotation degree $(t \cdot \Theta)$; $\mathcal{A}\in\mathbb{R}^{N\times N}$ is the token-wise attention score; $\operatorname{Mask}(\cdot)$ is a triangle causal mask; and $\tau$ is a learnable scalar parameter.

After multi-head concatenation and residual connection, let $\mathbf{u}_i\in\mathbb{R}^D$ denotes the embedding of the $i$-th token. The MoE module computes the output embeddings as follows: 
\begin{equation}
\begin{aligned}
\operatorname{MoE}\left({\mathbf{u}}_i\right) & = \sum_{j=1}^E\left(g_{j, i} \operatorname{FFN}_j({\mathbf{u}}_i)\right), \\
g_{j, i} & = \begin{cases}a_{j, i}, & a_{j, i} \in \operatorname{Topk}\left(\left\{a_{j, i} \mid 1 \leq j \leq E\right\}, K\right) \\
0, & \text { otherwise,}\end{cases} \\
a_{j, i} & =\operatorname{Softmax}_{j}(\mathbf{W}_j {\mathbf{u}}_i),
\end{aligned}
\end{equation}
where $E, K$ denote the number of routed and activated experts, respectively; $\operatorname{FFN}_j(\cdot)$ denotes the $j$-th routed expert; $g_{j,i}$ is the gated value for the $j$-th expert; $\operatorname{Topk}(\cdot, K)$ denotes the set comprising $K$ highest scores; and $a_{j,i}$ is the token-to-expert affinity score given by $E$ trainable parameter matrices $\mathbf{W}_j: \mathbb{R}^{D} \mapsto \mathbb{R}, j = 1,\ldots, E$. 

We empirically observe that a sparse MoE configuration, i.e., a large total number of experts but only a few activated per token ($E=32, K=2$), delivers optimal performance in Timer-S1. This configuration aligns with the global heterogeneity of time series data yet local simplicity of patch-level patterns, thus allowing the model to scale to the billion-scale parameters while improve inference speed.

We employ an auxiliary loss for load balancing among experts, defined as:
\begin{equation}
\mathcal{L}_{\text{aux}}=E\sum_{j=1}^E f_jP_j,\ f_j=\frac{1}{KN}\sum_{i=1}^{N}\boldsymbol{1}\left(a_{j, i} \in \operatorname{Topk}(\{a_{j, i} \mid 1 \leq j \leq E\}, K)\right),\ P_j=\frac{1}{N}\sum_{i=1}^N a_{j,i},
\end{equation}
where $f_j$ and $P_j$ represent the fraction of tokens assigned to expert $j$ and the proportion of router probability allocated to it, respectively, while $\boldsymbol{1}$ denotes the indicator function.

By passing through $L$ TimeMoE blocks, we obtain $N$ token embeddings $\{\mathbf{h}^L_i\}$ that aggregate the contextual information from preceding tokens, which are then projected by a forecasting head $\operatorname{PacthProject}(\cdot)$ to predict the next patch, optimized by the following next-token prediction (NTP) objective:
\begin{equation}
\hat{\mathbf{x}}_{i+1}=\operatorname{PacthProject}(\mathbf{h}_i^L),\ \mathcal{L}_{\text{NTP}}=\sum_{i=1}^N\mathcal{L}_{\text{pred}}(\mathbf{x}_{i+1}, \hat{\mathbf{x}}_{i+1}).
\end{equation}

\paragraph{TimeSTP} Despite the dense token-wise representations extracted by the main blocks, only the last-token embedding $\mathbf{h}^L_N$ can be used to generate one-step future predictions during inference. This forces the model to make multi-step autoregressive rollings for long-term forecasting, where error accumulation is particularly severe. To circumvent this, prior work employs multi-token prediction~\cite{das2023decoder, liu2025sundial} (using an output patch size $P_{\text{out}}>P_{\text{in}}$), which reduces rolling steps but fails to respect the serial nature of time series forecasting.

Intuitively, long-term forecasting requires serial computations of predictions. Instead of outputting a large patch at once or relying on iterative autoregressive rollouts, the proposed TimeSTP block progressively reuses and refines embeddings from the preceding block while continually attending to the initial input series.

Specifically, we append a sequence of $H=\lceil F/P \rceil-1$ TimeSTP blocks to predict the next $H$ patch tokens. Each block (indexed by $j$) contains a projection layer $\mathbf{M}_j\in\mathbb{R}^{D\times 2D}$ and a TimeMoE block. It operates by first concatenating two inputs: the token embeddings $\mathbf{h}_i^{L+j-1}$ from the preceding block (for $j=1$, it is the output embeddings $\mathbf{h}_i^{L}$ of the main blocks) and the initial input patch embeddings $\mathbf{h}_i^0$: 
\begin{equation}\label{equ:concat}
\bar{\mathbf{h}}_i^{ L+j}=\mathbf{M}_j\cdot\operatorname{Concat}\left(\operatorname{RMSNorm}(\mathbf{h}_i^{L+j-1}), \operatorname{RMSNorm}(\mathbf{h}^0_{i})\right).
\end{equation}
After the projection, these embeddings are fed into the internal TimeMoE block, as in Equation~\ref{equ:timemoe}:
\begin{equation}
\mathbf{h}_i^{L+j}=\operatorname{TimeMoE}(\bar{\mathbf{h}}_i^{ L+j})
\end{equation}
The output embeddings from each TimeSTP block are projected by a shared forecasting head and become predicted patches. Critically, the depth index $j$ of a TimeSTP block determines the next-patch offset in the final predictions, i.e., $j$-th TimeSTP block generates forecasts for the time series patch shifted by $j+1$. We formulate this Serial-Token Prediction (STP) objective as follows:
\begin{equation}
\hat{\mathbf{x}}_{i+j+1}=\operatorname{PacthProject}(\mathbf{h}_i^{L+j}),\ \mathcal{L}_{\text{STP}}=\frac{1}{H}\sum_{j=1}^H\sum_{i=1}^N\mathcal{L}_{\text{pred}}(\mathbf{x}_{i+j+1}, \hat{\mathbf{x}}_{i+j+1}).
\end{equation}
During training, we apply per-patch (dense) supervision to maintain the model's flexibility across variable context lengths. For inference, multi-step predictions are obtained in a single forward pass by using the last-token embedding from the main blocks $\mathbf{h}^{L}_N$, along with the last-token embedding $\mathbf{h}^{L+j}_N$ from all TimeSTP blocks. Empirically, predictions that are projected from the last-token embeddings yield the best performance.

TimeSTP distinguishes itself from multi-token prediction objective used in large language models~\cite{stern2018blockwise, liu2024deepseek, gloeckle2024better} in two key aspects. First, TimeSTP does not refer to future time series during training, since future ground-truth values are also unavailable during inference. Ignoring this train-test gap can lead to amplified error accumulation. Second, TimeSTP blocks are retained after training, while related work~\cite{liu2024deepseek} discards auxiliary blocks once the model is trained. This allows Timer-S1 to eliminate the rolling process and generate multiple patches through a single forward pass. Besides, our implementation enables the model to adaptively determine the inference depth (i.e., how many TimeSTP blocks to execute) according to the required forecasting horizon, thereby avoiding truncated multi-token predictions and unnecessary computational overheads.

\subsection{Forecasting Head}
Timer-S1 employs a shared quantile forecasting head $\operatorname{PatchProject}: \mathbb{R}^{D}\mapsto\mathbb{R}^{Q\times P}$ to generate $Q$ quantile predictions, each corresponding to a future patch of length $P$. The $k$-th quantile forecast, denoted as  $\hat{\mathbf{x}}^{(k)}$, is de-normalized using the input statistics $\mu$ and $\sigma$ (Equation~\ref{equ:denorm}). Thus, the training loss $\mathcal{L}_{\text{pred}}$ is defined as:
\begin{equation}
\begin{aligned}
\mathcal{L}_{\text{pred}}(\mathbf{x}, \hat{\mathbf{x}}) &=\frac{1}{Q}\sum_{k=1}^Q\text{wQL}_{q_k}(\mathbf{x}, \hat{\mathbf{x}}^{(k)}),\\
\text{wQL}_{q}(\mathbf{x}, \hat{\mathbf{x}}) &=2\frac{\sum_{t=1}^P\mathbf{\rho}_{q}(x_t, \hat{x}_t)}{\sum_{t=1}^P|x_t|},\\
\rho_q(x, \hat{x}) &= \begin{cases}(1-q) \cdot(\hat{x}-x), & \text { if } x<\hat{x}, \\ q \cdot(x-\hat{x}), & \text { if } x \geq \hat{x},\end{cases}
\end{aligned}
\end{equation}
where wQL, the weighted mean of Quantile Loss~\cite{aksugift}, is a common approximation of the Continuous Ranked Probability Score~\cite{gneiting2007strictly} (CRPS) commonly adopted in probabilistic forecasting. Here, $\rho_q(\cdot, \cdot)$ denotes the pinball loss~\cite{steinwart2011estimating}, which evaluates the prediction accuracy at the quantile level $q$.

Timer-S1 adopts the quantile set $q_k\in\{0.1, 0.2,\dots,0.9\}$, following the task configuration of the GIFT-Eval leaderboard. It is worth noting that the architecture of Timer-S1 is general to leverage other forecasting heads, such as linear projection, parametric probabilistic heads~\cite{salinas2020deepar}, or diffusion-based heads~\cite{liu2025sundial}. Furthermore, both the patch embedding and patch projection layers are shared across all TimeMoE and TimeSTP blocks. This design ensures consistent transformation from token embeddings to time series patches and promotes parameter efficiency through dense supervision of these shared components.

\section{Training}\label{sec:train}
\begin{figure*}[htb]
\centering
\includegraphics[width=\textwidth]{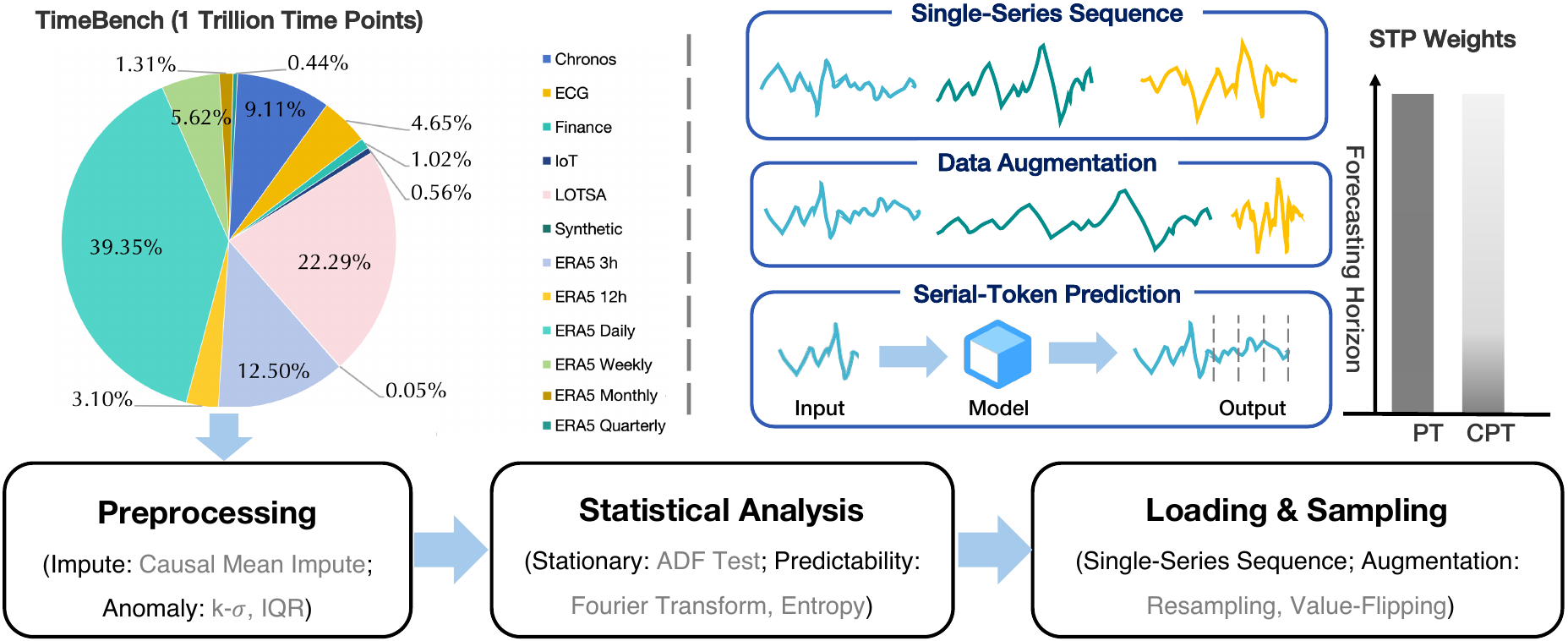}
\caption{Illustration of the TimeBench dataset and the training pipeline of Timer-S1. TimeBench integrates over one trillion time points from multiple domains, processed through quality-focused preprocessing, predictability assessment, and diversity-enhancing augmentation. TimeBench is loaded in a single-series sequence format for learning univariate patterns. Timer-S1’s training follows a multi-stage design: it is first pre-trained via serial-token prediction with uniform horizon weighting; then it undergoes continued pre-training using a horizon-decay objective to enhance short-term accuracy; and extends its context length from 2880 to 11520 through long-context adaptation.} 
\label{fig:training}
\end{figure*}

In this section, we introduce data and training details of Timer-S1 (Figure~\ref{fig:training}). We begin with data curation, presenting TimeBench, a high-quality dataset with one trillion time points. Specifically, we elaborate on data augmentation aimed at mitigating predictive bias. To cope with the heterogeneity among datasets, we pre-train Timer-S1 with the single-sequence series format, using the serial-token prediction objective. Finally, we conduct post-training to enhance short-term and long-context capabilities of the pre-trained model.

\subsection{Data}
The success of foundation models relies heavily on large-scale, high-quality datasets. Time series data, while ubiquitous, often present curation challenges, including missing values, unpredictable dynamics, and structural variance, all of which have an adverse effect on learning. To overcome these issues, we develop a comprehensive data curation pipeline that data preprocessing, statistical analysis, and loading samples. Notably, all instances that could potentially lead to test data leakage in GIFT-Eval~\cite{aksugift} are carefully removed.

\paragraph{TimeBench} To enhance data diversity, we collect real-world time series from common domains such as finance, IoT, meteorology, and healthcare, and incorporate publicly released time series from research efforts including Chronos~\cite{ansari2024chronos} and LOTSA~\cite{woo2024unified}. We select variates guided by a proxy criterion, where a variate with a strong autoregressive property is selected, using the statistical significance of fitted ARIMA models.

To enrich pattern variety, we leverage synthetic data, including canonical signals, e.g., linear, sinusoidal, exponential, power, impulse, and step functions, as well as their additive and multiplicative combinations. We also adopt KernelSynth~\cite{ansari2024chronos} to sample from random-instantiated temporal causal models.

To ensure high data quality, we perform rigorous preprocessing, including causal mean imputation, outlier removal based on k‑$\sigma$ and IQR thresholds using a shifting window. We preserve original timestamps; for data without timestamps, we assign default numeric indices starting from 0. The final curated corpus, TimeBench contains 1032 billion regularly sampled time points. Detailed proportions are provided in Figure~\ref{fig:training}.

Finally, each dataset is assessed using two key metrics: (1) Augmented Dickey–Fuller (ADF)~\cite{adftest} test statistic, and (2) a forecastability measure~\cite{goerg2013forecastable} based on the entropy of spectral components, defined as follows:

\begin{equation}
\begin{aligned}
\operatorname{ADF-Statistic}(\mathcal{D})& =\sum_{i=1}^C \frac{T_i}{T} \operatorname{ADF-Statistic}\left(\mathbf{S}^{(i)}\right), \\
\text {Forecastability}(\mathcal{D}) & =\sum_{i=1}^C \frac{T_i}{T}\left(1-\operatorname{Entropy}\left(\mathcal{F}\left(\mathbf{S}^{(i)}\right)\right)\right),
\end{aligned}
\end{equation}

where $\mathbf{S}_i \in \mathbb{R}^{T_i}$ denotes the $i$-variate in the dataset $\mathcal{D}$; $C$ is the variate number; $T_i$ is the corresponding length; $T=\sum_{i=1}^C T_i$ is the total length of the dataset; and $\mathcal{F}(\mathbf{S}^{(i)})$ denotes the Fourier decomposition of the variate. The above two metrics serve as coordinates defining a two-dimensional “complexity plane” for each dataset.

\paragraph{Data Augmentation} Enhancing the diversity of time series data has long been a challenge in building time series foundation models. However, real-world time series often follow imbalanced distributions, leading to predictive bias in the trained model. The model may present varied performance on data with different frequencies, or it tends to produce predictions with a stereotypical tendency. Therefore, we delve into data augmentation and identify several effective techniques that mitigate such predictive bias.

\begin{itemize}
    \item \textbf{Resampling} We vary the sampling rate of the original series through down‑sampling and interpolation on Fourier bases to expose the model to diverse temporal resolutions. This encourages robustness to frequency shifts and improves generalization across different sampling regimes encountered in practice.
    \item \textbf{Value‑Flipping} The input and output series are both multiplied by –1, thereby inverting their trends while preserving temporal dependencies. This simple operation counteracts the model’s tendency to latch onto persistent directional trends, a predictive bias observed in our previous work~\cite{liu2025sundial}.
\end{itemize}

\subsection{Pre-Training} Timer-S1 is configured with a hidden dimension $D=1024$, patch size $P=16$, and an input token number of $N=180$, using $L=24$ TimeMoE as the main blocks and $H=16$ TimeSTP blocks. It leads to a maximum context length $T=2880$ and prediction length $(H+1)P=272$ in a single inference pass. Timer-S1 adopts a sparse Mixture-of-Experts configuration ($E=32$ and $K=2$). The pre-training objective is formulated as:
\begin{equation}
\mathcal{L}_{\text{Pre-train}}=\mathcal{L}_{\text{NTP}}+\mathcal{L}_{\text{STP}}+\alpha\mathcal{L}_{\text{aux}},
\end{equation}
where $\alpha$ is the hyperparameter that balances the MoE auxiliary loss. We assign equal weights to the next-token and serial-token prediction objectives, and each TimeSTP block is assigned an equal weight for different horizons. It means we construct a dense set of forecasting tasks from each time series, where arbitrary lengths can serve as input or output. By treating all such tasks equally, the first-stage pre-training achieves two goals: (1) it maximizes sample efficiency from the raw series and (2) ensures the TimeMoE module (for contextual representation) and the TimeSTP module (for multi-patch prediction) are fully trained.

\subsection{Post-Training} 
Considering the serial nature of time series forecasting, where each prediction depends on all prior estimations, long-term forecast accuracy relies fundamentally on short-term performance. To enhance this capability, our post-training stage focuses exclusively on short-term forecasting. First, we adopt the datasets for short-term tasks in GIFT-Eval Pretrain~\cite{aksugift} for post-training. Next, we propose a weighted STP objective that prioritizes short-term forecasting performance and a data revisiting mechanism to avoid overfitting. We also perform context length extension to equip the model with sufficient historical information for more precise predictions.

\paragraph{Weighted Serial-Token Prediction} In Timer-S1, each TimeSTP block is specialized for a specific forecasting horizon. While STP is tailored to improve long-term forecasting performance, short-term forecasting, as the initial step of long-term forecasting, should be further enhanced. Therefore, we conduct continued pre-training with a weighted Serial-Token Prediction (wSTP) loss that prioritizes the learning of shallow STP blocks:
\begin{equation}\label{equ:wstp}
\begin{aligned}
\mathcal{L}_{\text{Post-train}}&=\mathcal{L}_{\text{NTP}}+\mathcal{L}_{\text{wSTP}}+\alpha\mathcal{L}_{\text{aux}},\\
\mathcal{L}_{\text{wSTP}}&=\frac{1}{H}\sum_{j=1}^H \frac{1}{\sqrt{j}}\sum_{i=1}^N\mathcal{L}_{\text{pred}}(\mathbf{x}_{i+j+1}, \hat{\mathbf{x}}_{i+j+1}).
\end{aligned}
\end{equation}
To mirror the increasing uncertainty in long-term forecasts, we apply a weight decay of $\frac{1}{\sqrt{j}}$ on deeper TimeSTP blocks. This decay rate is derived from the linear growth of variance in a standard first-order Markov process.

\paragraph{Continued Pre-Training} At the post-training stage, we propose a data revisiting mechanism, where training data is sampled from the mixture of GIFT-Eval Pretrain and TimeBench. This approach mitigates overfitting to the distribution of the post-training dataset and enhances generalization across other data. At this stage, we also perform a direct context extension via RoPE~\cite{su2024roformer}, extending the context length from $2880$ to $11520$.

\subsection{Training Infrastructure}
The training of Timer-S1 is supported by VeOmni~\cite{ma2025veomni}, a unified framework designed for pre-training and post-training foundation models, which enables seamless scaling of Timer-S1 to billion-scale parameters across multiple devices. To ensure training efficiency, Timer-S1 adopts BF16 precision.

Raw data in TimeBench is converted into compressed Parquet files containing both values and timestamps, resulting in approximately $4$TB of physical storage. Sequences are sampled via an in-memory sliding window, which enables random access to time series but may lead to data inflation. To address this problem, we implement a hybrid memory‑disk loading strategy. Specifically, we partition the TimeBench dataset into $50$MB shards, a size chosen to balance I/O concurrency and sampling randomness. Each shard serves as the basic unit for in‑memory sliding-window sampling, and an in‑memory queue is maintained to manage active shards and avoid loading entire sequences from TimeBench at once.

\section{Experiments}\label{sec:exp}
In this section, we conduct comprehensive evaluations. We assess Timer-S1's performance on GIFT-Eval~\cite{aksugift} against state-of-the-art models, tracking improvements of the post-training stage. We validate the efficacy of serial-token prediction by comparing it with next-token prediction~\cite{bengio2000neural, liutimer} and multi-token prediction~\cite{das2023decoder, liu2025sundial}. We analyze scaling behavior by comparing various model configurations. We conduct detailed ablation studies on key components, including TimeSTP variants, data augmentation, and pre-training effect.

\begin{figure*}[hbt]
\centering
\includegraphics[width=\textwidth]{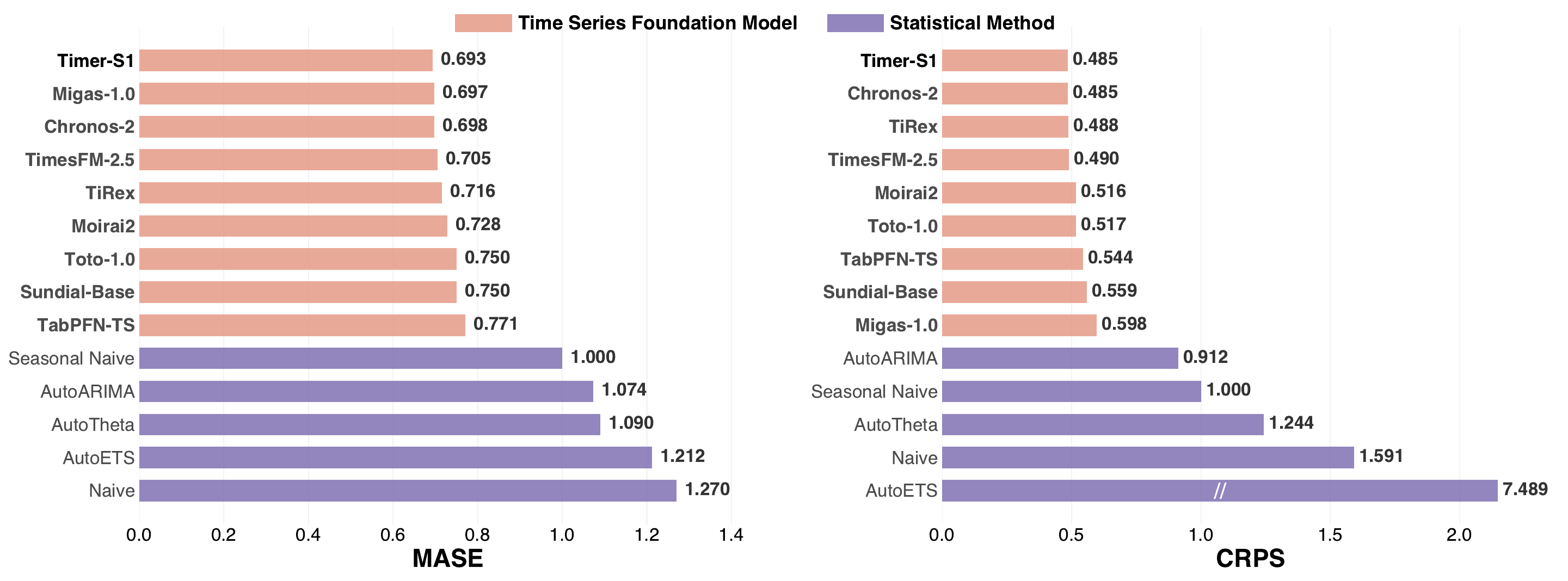}
\caption{Performance of Timer-S1 on the GIFT-Eval leaderboard.} 
\label{fig:gift_eval}
\end{figure*}

\subsection{Benchmark Results}
We follow the official evaluation protocol of GIFT-Eval, which comprises 24 datasets spanning 144,000 time series and 177 million data points, using Mean Absolute Scaled Error (MASE) for point forecasting and Continuous Ranked Probability Score (CRPS) for probabilistic forecasting. This benchmark has included well-established baseline models, covering advanced time series foundation models and statistical methods.

\begin{figure*}[ht]
\centering
\includegraphics[width=\textwidth]{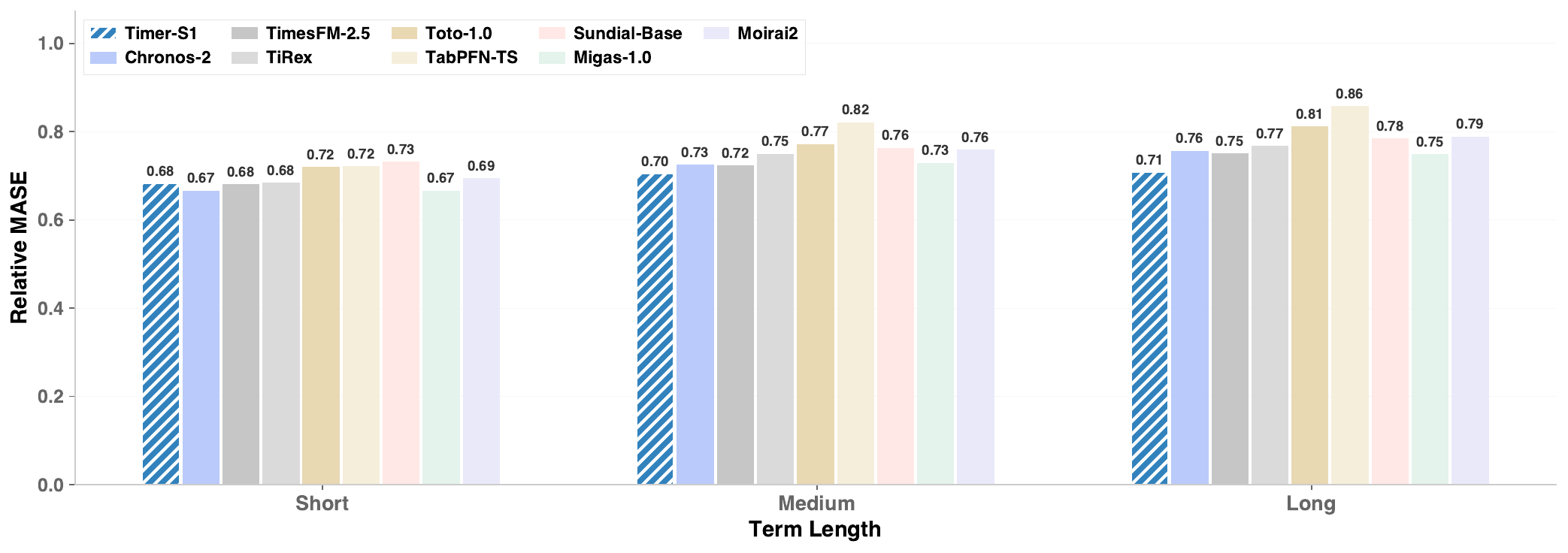}
\caption{Performance (MASE) on the GIFT-Eval leaderboard, grouped by the term length.} 
\label{fig:gift_eval_term_mase}
\end{figure*}

\begin{figure*}[ht]
\centering
\includegraphics[width=\textwidth]{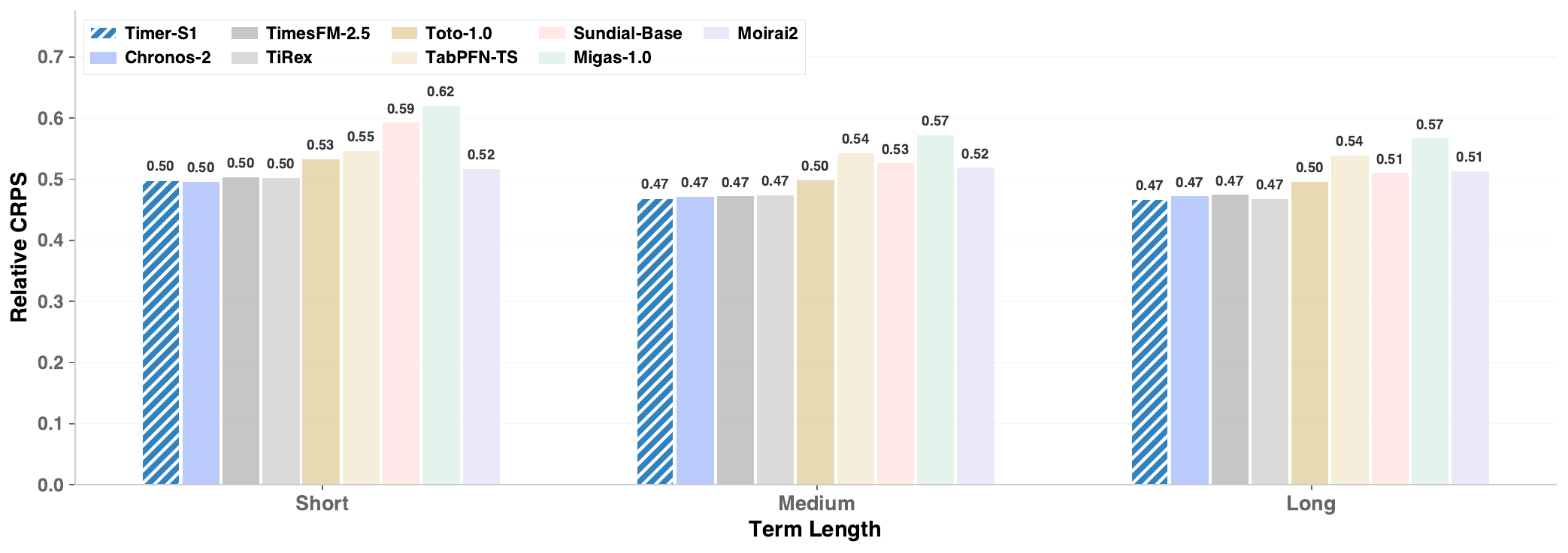}
\caption{Performance (CRPS) on the GIFT-Eval leaderboard, grouped by the term length.} 
\label{fig:gift_eval_term_crps}
\end{figure*}
\vspace{-5pt}

\paragraph{GIFT-Eval} As shown in Figure~\ref{fig:gift_eval}, Timer-S1 achieves state-of-the-art MASE and CRPS. While Timer-S1 does not explicitly utilize multivariate interactions like Chronos-2~\cite{ansari2025chronos}, it attains competitive results. Trained on the same TimeBench, Timer-S1 shows a 7.6\% lower MASE and a 13.2\% lower CRPS than Timer-3 (Sundial), validating the serial scaling effect of our foundation model. The evaluation presents the competitiveness of Timer-S1 as a zero-shot forecaster and a base model in future agentic forecasting systems. 

To better understand the origin of Timer-S1's performance advantage, we analyze the results grouped by the forecasting term length of GIFT-Eval (three prediction settings based on frequency and domain). As shown in Figure~\ref{fig:gift_eval_term_mase}-\ref{fig:gift_eval_term_crps}, we clearly observe that Timer-S1 achieves substantially better performance on the medium- and long-term tasks. This reinforces the effectiveness of our serial forecasting approach, which improves the performance on challenging long-term forecasting tasks through crucial serial computations.

\begin{figure*}[htb]
\centering
\includegraphics[width=\textwidth]{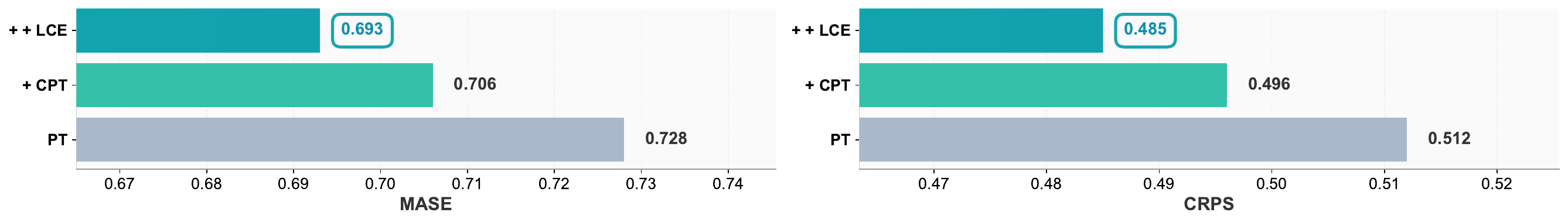}
\caption{Performance of Timer-S1 with different training pipelines on the GIFT-Eval leaderboard.} 
\label{fig:multi_stage}
\end{figure*}

\paragraph{Post-Training} As shown in Figure~\ref{fig:multi_stage}, the performance is improved with continued pre-training and long-context extension. The performance gain over a single pre-training stage once again confirms the technical route that foundation models require multi-stage training and distinct objectives at different phases~\cite{guo2025deepseek}. Similarly, Chronos-2 is pre-trained with a small number of output patches and then post-trained with an increased number. We implement an opposite pipeline: the first stage is dedicated to fully training TimeSTP across all horizons, whereas the post-training stage shifts focus to short-term tasks via a weighted objective.

\subsection{Scaling Analysis}
In this section, we compare serial-token prediction against next-token prediction and multi-token prediction, verifying that serial forecasting helps to scale time series foundation models and results in better performance. We compare different backbone configurations in the pre-training stage, confirming an appropriate model size. 

\begin{figure*}[htb]
\centering
\includegraphics[width=\textwidth]{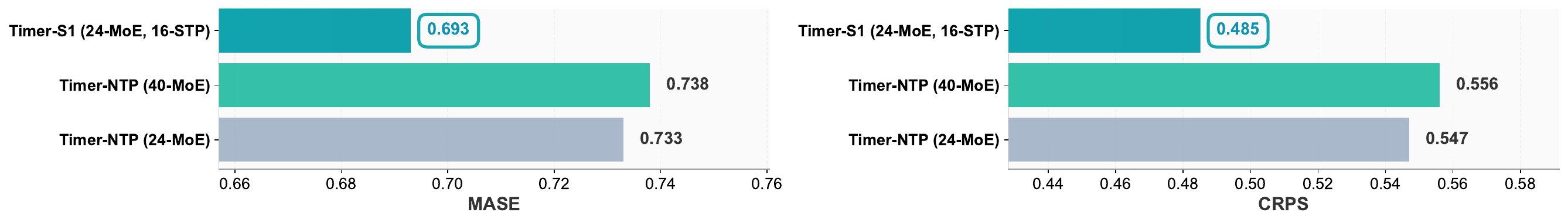}
\caption{Performance of the Timer model trained by next-token prediction and serial-token prediction.} 
\label{fig:compare_token_prediction_16-16}
\end{figure*}

\begin{figure*}[htb]
\centering
\includegraphics[width=\textwidth]{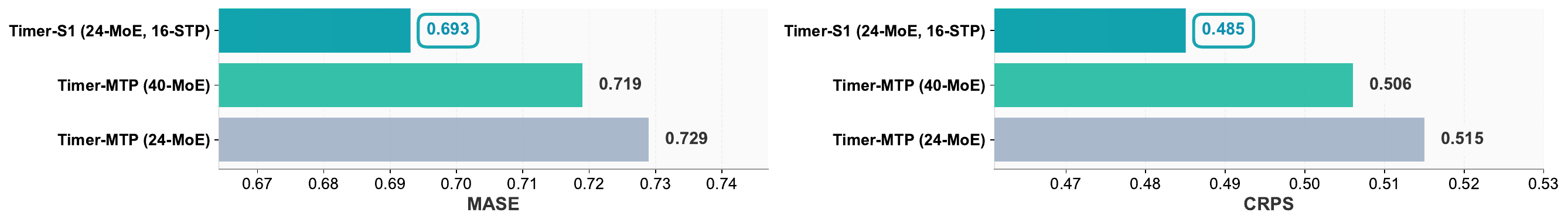}
\caption{Performance of the Timer model trained by multi-token prediction and serial-token prediction.} 
\label{fig:compare_token_prediction_16-272}
\end{figure*}

\begin{figure*}[htb]
\centering
\includegraphics[width=\textwidth]{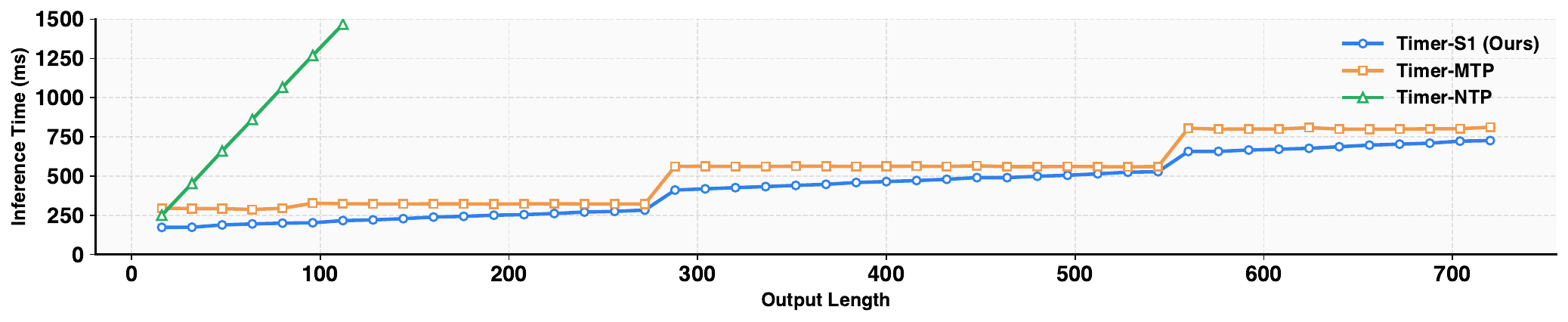}
\caption{Inference time of models trained by next-token prediction, multi-token prediction, and serial-token prediction. We set the same  backbone configuration except for the training objective. The input sequence length is set to 11520.} 
\label{fig:infer_time_comparison_ms_p50}
\end{figure*}

\paragraph{Serial-Token Prediction} We compare models trained by different objectives in Figure~\ref{fig:compare_token_prediction_16-16}-\ref{fig:compare_token_prediction_16-272} (the backbone configuration is provided). Different from next-token prediction (NTP), serial-token prediction (STP) explicitly reduces rolling iterations, thereby mitigating the effect of error accumulation. Multi-token prediction (MTP) produces outcomes of all horizons in a single forward pass, but it lacks the serial computations necessary for long-term forecasting, which becomes a bottleneck in achieving higher precision. Notably, Timer-S1 (24-MoE, 16-STP) achieves better results than Timer-NTP (40-MoE) and Timer-MTP (40-MoE) under the same budget of block numbers, indicating that TimeSTP can be a generic module for improving long-term forecasting performance. We also compare the inference time in Figure~\ref{fig:infer_time_comparison_ms_p50}. To produce the next prediction, Timer-NTP is required to pass through the whole model, while Timer-S1 only needs to pass by a single TimeSTP block. In a single inference pass (output length $= 272$ in per roll), Timer-MTP adopts a larger forecasting head and needs to truncate redundant predictions, leading to additional computations.

\begin{figure*}[htb]
\centering
\includegraphics[width=\textwidth]{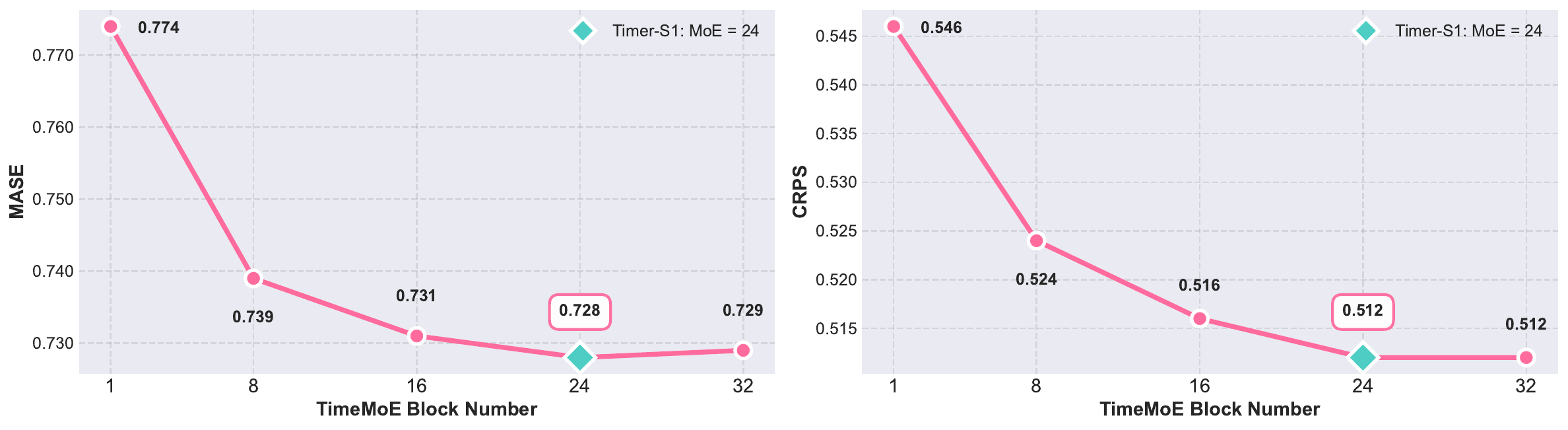}
\caption{Performance of the pre-trained Timer-S1 with varying TimeMoE blocks (fixed 16-block TimeSTP).} 
\label{fig:main_layer_scaling_experiment}
\end{figure*}

\begin{figure*}[htb]
\centering
\includegraphics[width=\textwidth]{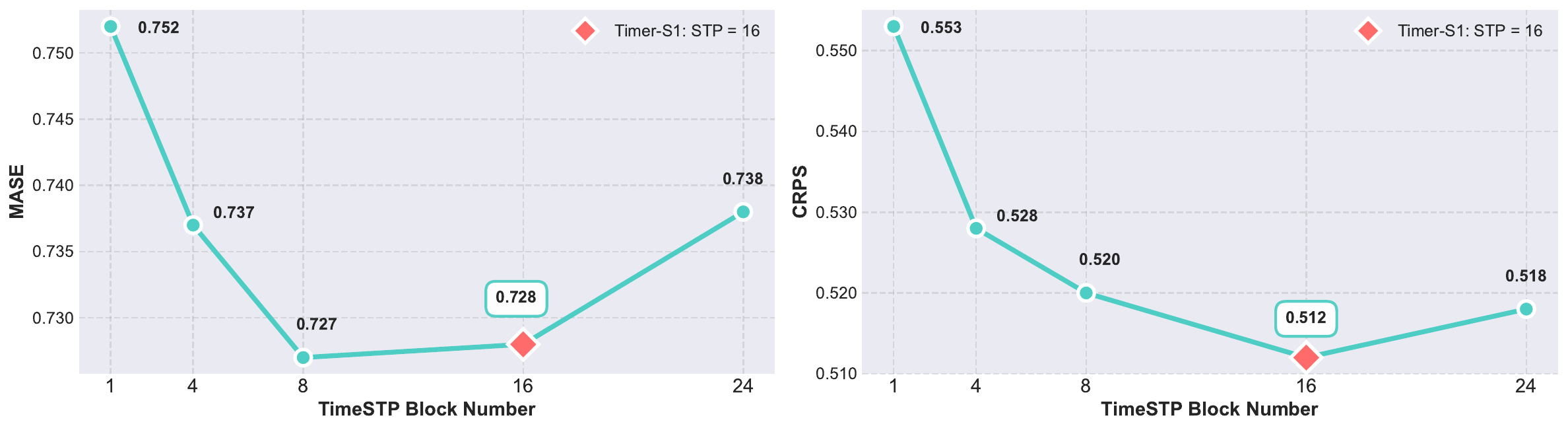}
\caption{Performance of the pre-trained Timer-S1 with varying TimeSTP blocks (fixed 24-block TimeMoE).} 
\label{fig:stp_scaling_experiment}
\end{figure*}

\paragraph{Model Configuration} During the pre-training stage, we change the backbone configuration in TimeMoE (responsible for extracting contextual representations) and TimeSTP (which performs sequential forecasting) numbers to investigate the scaling behavior (Figures~\ref{fig:main_layer_scaling_experiment} and~\ref{fig:stp_scaling_experiment}). The model continues to benefit from scaling up to the billion level comprising 24 TimeMoE blocks and 16 TimeSTP blocks, which surpasses the parameter scale of existing time series foundation models and validate the scaling law.

\subsection{Ablation Study}
In this section, we evaluate the variants of TimeSTP, demonstrating that the current design is specialized for the evolving property of time series data. We evaluate our data augmentation method. We compare Timer-S1 with the model trained from scratch, which implies knowledge transfer in our pre-trained models.

\paragraph{TimeSTP Design} Serial forecasting contains serial computations across blocks: longer-horizon predictions are processed through more Transformer blocks. Each TimeSTP block produces embeddings that serve two purposes: (1) they are projected by the forecasting head to generate predictions for the current horizon, and (2) they are passed to the next TimeSTP block and fused with the original input embeddings. In contrast, a similiar implementation in LLMs~\cite{liu2024deepseek} utilizes shifted embeddings from future inputs, which are available only during training, and discards the auxiliary prediction blocks afterward.

\begin{figure*}[htbp]
\centering
\includegraphics[width=\textwidth]{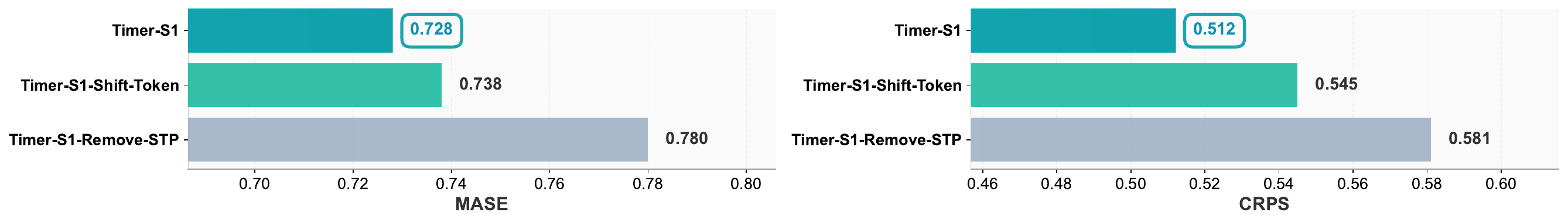}
\caption{Performance of Timer-S1 with TimeSTP variants.} 
\label{fig:model_comparison_STP}
\end{figure*}

We compare this variant (Timer-S1-Shift-Token) in Figure~\ref{fig:model_comparison_STP}, which leads to worse performance compared to non-shifting design. The main reason stems from the distributional variability of time series, where the train-test gap has a pronounced impact on time series forecasting. Furthermore, to keep the consistency of model structure between training and inference, the trained TimeSTP blocks are also reserved as an inference component in Timer-S1. We evaluate a variant choice (Timer-S1-Remove-STP) in Figure~\ref{fig:model_comparison_STP}, where discarding trained TimeSTP and adopting rolling-style autoregressive forecasting lead to significantly worse performance.

\begin{figure*}[ht]
\centering
\includegraphics[width=\textwidth]{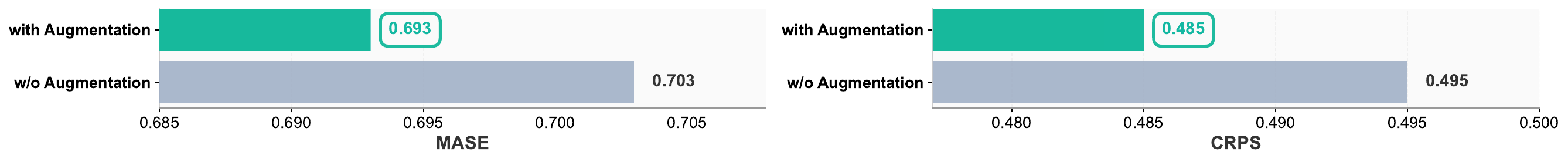}
\caption{Performance of Timer-S1 with and without data augmentation.} 
\label{fig:model_comparison_augmentation}
\end{figure*}

\begin{figure*}[ht]
\centering
\includegraphics[width=\textwidth]{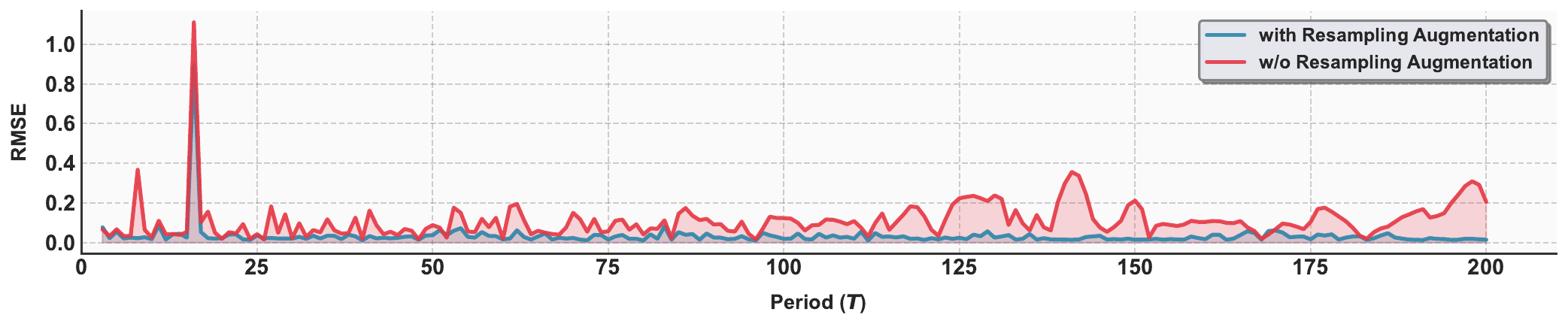}
\caption{Performance of of Timer-S1 on different sinusoidal signals.} 
\label{fig:comparison_resampling}
\end{figure*}

\paragraph{Data Augmentation}\label{sec:data_aug} Figure~\ref{fig:model_comparison_augmentation} illustrates the performance gains from data augmentation. To further elucidate the effect, we compare forecasting cases on sinusoidal signals (Figure~\ref{fig:comparison_resampling}). The resampling augmentation helps the model to generalize across different frequencies, improving the robustness to temporal resolution shifts. Notably, an error spike emerges at a period of approximately 16, which is exactly the configured fixed patch size, highlighting a promising direction for further model refinement.

\begin{figure*}[ht]
\centering
\includegraphics[width=\textwidth]{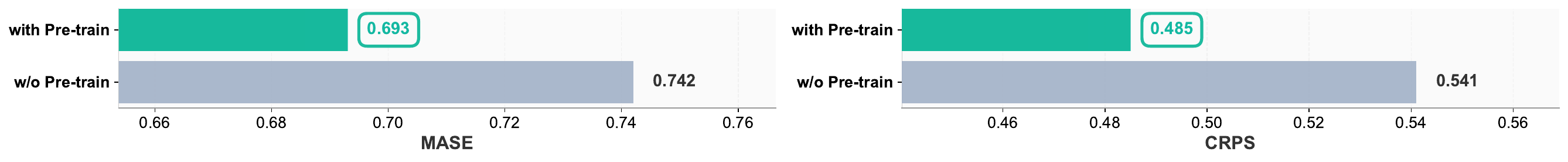}
\caption{Performance of Timer-S1 with and without the pre-training on TimeBench.} 
\label{fig:comparison_ft}
\end{figure*}

\paragraph{Pre-training on TimeBench} We compare the performance of Timer-S1 with and without the pre-training on TimeBench in Figure~\ref{fig:comparison_ft}. Based on the same model configuration and post-training dataset, the model trained from scratch has significantly worse results on the GIFT-Eval leaderboard. This confirms the generalization of temporal patterns learned during pre-training. Even though Timer-S1 focuses solely on univariate context, the pre-trained model can be effectively transferred to a variety of downstream tasks.

\section{Conclusion}\label{sec:conclusion}
In this technical report, we introduce Timer-S1, a billion-scale Mixture-of-Experts (MoE) time series foundation model that addresses scalability bottlenecks through Serial Scaling. The core innovation of the model lies in Serial-Token Prediction (STP), a generic objective that respects the serial nature of forecasting, improving long-term forecasting performance with gradually increased serial computations. By pre-training on TimeBench of over one trillion points, Timer-S1 benefits from data augmentation and a multi-stage training pipeline. The outcome model yields state-of-the-art performance on the GIFT-Eval leaderboard.

Although Timer-S1 presents strong forecasting results, it has certain limitations. First, it does not natively incorporate exogenous covariates, which leaves large room for improvement in the future. This capability gap largely stems from the training difficulty on unstructured multivariate datasets. To address this, we plan to expand the synthesis of multivariate data and upgrade the pre-training framework. Second, considering the fundamental differences between short-term and long-term forecasting tasks, it is crucial to develop an adaptive representation learning paradigm that improves the model’s generality across varying input contexts and output horizons. Finally, we will integrate Timer-S1’s general forecasting capabilities into agentic systems for making multimodal forecasting, reasoning, and planning.

\clearpage

\bibliographystyle{plainnat}
\bibliography{main}


\end{document}